\documentclass[10pt, a4paper]{article}

\usepackage{lrec2026}

\usepackage{graphicx}
\usepackage{amsmath}
\usepackage{multirow}
\usepackage{tabularx}
\usepackage{booktabs}
\usepackage{pifont}
\usepackage{amssymb}
\usepackage{enumitem}
\usepackage[table,xcdraw]{xcolor}
\usepackage{subcaption}
\usepackage{cleveref}
\usepackage{tcolorbox}
\usepackage[absolute,overlay]{textpos}
\usepackage{cuted}
\usepackage{caption}

\crefname{footnote}{Footnote}{Footnotes}
\newcommand{\footnoteref}[1]{\hyperref[#1]{Footnote~\ref*{#1}}}

\title{LIT-RAGBench: Benchmarking Generator Capabilities of \\Large Language Models in Retrieval-Augmented Generation}

\name{
\begin{tabular}{c}
Koki Itai$^{1,2}$, Shunichi Hasegawa$^{1}$, Yuta Yamamoto$^{1}$, \\
Gouki Minegishi$^{1,3}$, Masaki Otsuki$^{1,3}$
\end{tabular}
}

\address{
$^{1}$neoAI Inc., Tokyo, Japan \\
$^{2}$Tokyo Metropolitan University, Tokyo, Japan \\
$^{3}$The University of Tokyo, Tokyo, Japan \\
\{k.itai, s.hasegawa, y.yamamoto, g.minegishi, m.otsuki\}@neoai.jp
}

\abstract{
Retrieval-Augmented Generation (RAG) is a framework in which a Generator, such as a Large Language Model (LLM), produces answers by retrieving documents from an external collection using a Retriever.
In practice, Generators must integrate evidence from long contexts, perform multi-step reasoning, interpret tables, and abstain when evidence is missing. 
However, existing benchmarks for Generators provide limited coverage, with none enabling simultaneous evaluation of multiple capabilities under unified conditions.
To bridge the gap between existing evaluations and practical use, we introduce LIT-RAGBench (the Logic, Integration, Table, Reasoning, and Abstention RAG Generator Benchmark), which defines five categories: Integration, Reasoning, Logic, Table, and Abstention—each further divided into practical evaluation aspects.
LIT-RAGBench systematically covers patterns combining multiple aspects across categories. 
By using fictional entities and scenarios, LIT-RAGBench evaluates answers grounded in the provided external documents.
The dataset consists of 114 human-constructed Japanese questions and an English version generated by machine translation with human curation.
We use LLM-as-a-Judge for scoring and report category-wise and overall accuracy. 
Across API-based and open-weight models, no model exceeds 90\% overall accuracy.
By making strengths and weaknesses measurable within each category, LIT-RAGBench serves as a valuable metric for model selection in practical RAG deployments and for building RAG-specialized models.\\
\newline  
}

\begin{document}
\begin{textblock*}{10cm}(2cm,1cm)
	\underline{Published as a conference paper at LREC 2026}
\end{textblock*}

\maketitleabstract

\begin{strip}
\centering
\includegraphics[width=\textwidth]{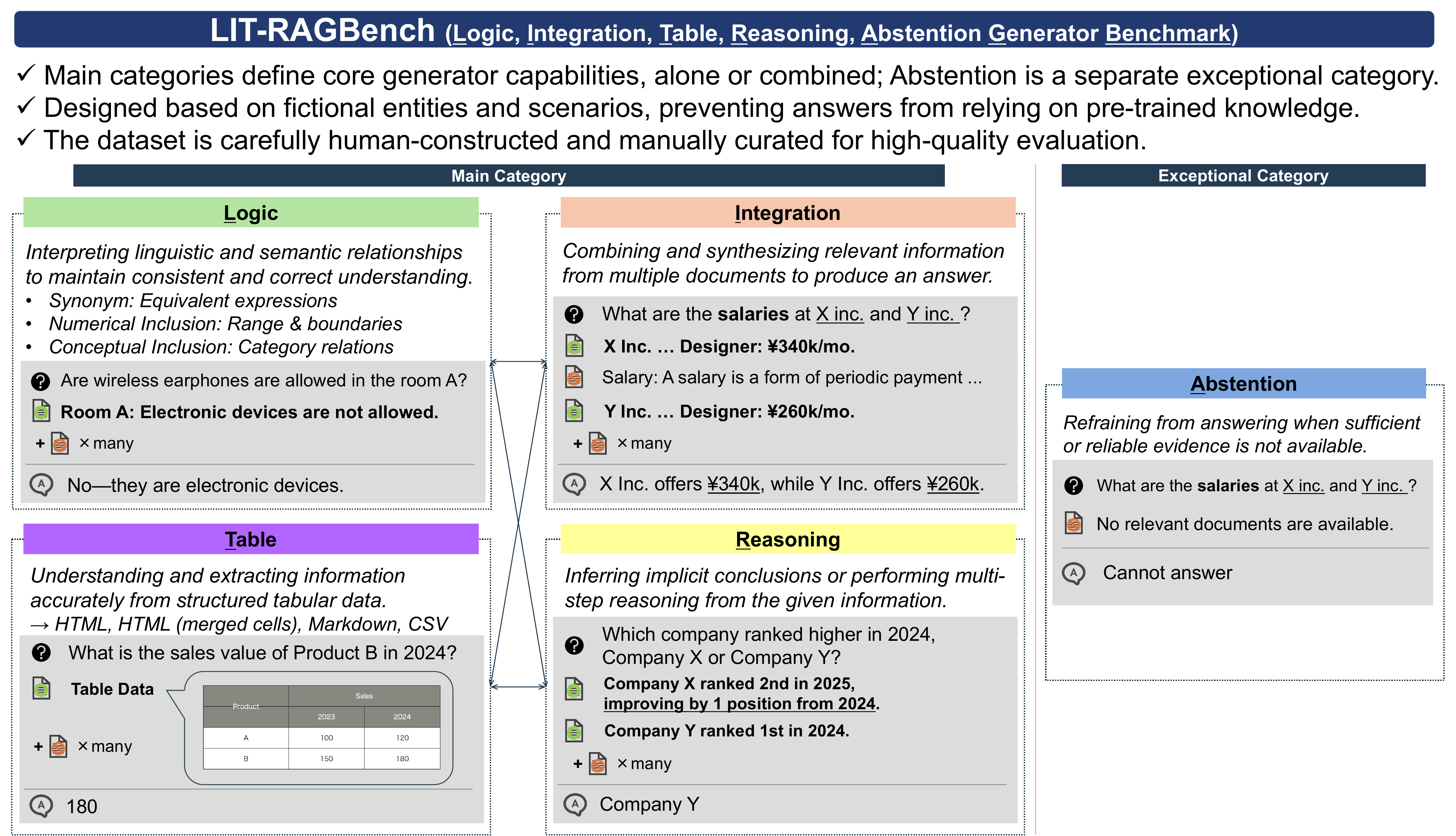}
\captionof{figure}{Illustration of the evaluation categories of LIT-RAGBench. These categories reflect the capabilities required of the Generator in RAG based on real-world scenarios.}
\label{fig:proposal}
\end{strip}

% \begin{figure*}[t]
%     \begin{center}
%     \includegraphics[width=\textwidth]{figures/key_visual_20260429.pdf}
%     \caption{Illustration of the evaluation categories of LIT-RAGBench. These categories reflect the capabilities required of the Generator in RAG based on real-world scenarios.}
%     \label{fig:proposal}
%     \end{center}
%     % \vspace{-2mm}
% \end{figure*}

%%%%%%%%%%%%%%%%%%%%%%%%%%%%%%%%%%%%%%%%%%%%%%%%%%%%%%%%%%
%%%%%%%%%%%%%%%%%%%%%  Introduction  %%%%%%%%%%%%%%%%%%%%%
%%%%%%%%%%%%%%%%%%%%%%%%%%%%%%%%%%%%%%%%%%%%%%%%%%%%%%%%%%

\section{Introduction}
Recent advancements in Large Language Models (LLMs) have significantly enhanced their capabilities across multiple domains~\citep{brown2020_fewshotlearners, openai2024gpt4technicalreport, LLM_Survey2024}.
However, several challenges have been reported, including factually ungrounded hallucinations~\citep{cao-etal-2020-factual, Hallucination_survey}, outdated information~\citep{he2022rethinkingretrievalfaithfullarge}, and limited domain-specific expertise~\citep{Financial_tasks, Hallucination_survey,outdated_knowledge}.
Retrieval-Augmented Generation (RAG) has emerged as a valuable approach to address these challenges~\citep{rag_original_paper, RAG_Survey}.
RAG is a framework in which a Generator, such as an LLM, produces answers based on documents retrieved by a Retriever from an external collection.
In practical applications, the Generator is required to accurately extract evidence from context while demonstrating multifaceted abilities, such as referencing and integrating evidence from multiple documents, performing multi-hop reasoning, and interpreting tabular data.
Although many benchmarks~\citep{rgb, frames} have been proposed to evaluate the Generator, they do not adequately cover the diverse capabilities needed in real-world RAG scenarios. Moreover, practical scenarios often require multiple capabilities simultaneously, yet no existing benchmark systematically evaluates such combinations under unified conditions.

% \begin{figure}[t]
%     \begin{center}
%     \includegraphics[width=\columnwidth]{figures/LIT-RAGBench.pdf}
%     \caption{Illustration of the evaluation categories of LIT-RAGBench. These categories reflect the capabilities required of the Generator in RAG based on real-world scenarios.}
%     \label{fig:proposal}
%     \end{center}
%     % \vspace{-2mm}
% \end{figure}

To bridge the gap between existing evaluations and practical use, this study proposes LIT-RAGBench (the Logic, Integration, Table, Reasoning, and Abstention RAG Generator Benchmark), designed to evaluate the Generator independently of retrieval quality.
LIT-RAGBench defines fundamental Generator capabilities as ``evaluation categories'', each comprising detailed ``evaluation aspects'' derived from real-world RAG use cases. 

The overall structure of these categories is illustrated in \autoref{fig:proposal}. Specifically, LIT-RAGBench evaluates five core aspects of Generator capability: (1) \emph{Integration}: generating information from multiple sources; (2) \emph{Reasoning}: inferring implicit conclusions from retrieved information; (3) \emph{Logic}: maintaining semantic and deductive consistency; (4) \emph{Table}: comprehending and interpreting tabular data; and (5) \emph{Abstention}: refraining from answering when reliable evidence cannot be established.
The evaluation datasets were constructed through a hybrid approach that combines LLM-based synthetic data generation and human curation, followed by manual filtering to ensure quality.

Evaluation experiments on major LLMs, including both API-based and open-weight models, revealed distinct performance patterns across categories.
No model exceeded 90\% overall accuracy, with performance variations across categories revealing each model's strengths and weaknesses. 
These findings demonstrate that LIT-RAGBench serves as a useful metric for model selection in practical RAG deployment and for building RAG-specialized models.
To facilitate reproducibility and further research, we release the dataset, the LLM prompts used for dataset construction and evaluation, and the corresponding code\footnote{\label{footnote_repo}\url{https://github.com/Koki-Itai/LIT-RAGBench}}.

%%%%%%%%%%%%%%%%%%%%%%%%%%%%%%%%%%%%%%%%%%%%%%%%%%%%%%%%%%
%%%%%%%%%%%%%%%%%%%%%  Preliminaries  %%%%%%%%%%%%%%%%%%%%%%%
%%%%%%%%%%%%%%%%%%%%%%%%%%%%%%%%%%%%%%%%%%%%%%%%%%%%%%%%%%
\section{Preliminaries}
This section formalizes the RAG process for subsequent explanation.
Let $\mathcal{R}$ and $\mathcal{G}$ denote the Retriever and Generator components, respectively.
$\mathcal{R}$ takes a query $x_{r}$ as input and outputs a set of related text segments $C=\{c_1,\ldots,c_n\}$ from an external data source $\mathcal{E}$ such as a database or the Web. 
$c$ is a text segment called a chunk, which is created by splitting documents stored in $\mathcal{E}$ into shorter segments for search efficiency.
The value $n$ is the number of chunks retrieved by $\mathcal{R}$, specified by the developer. 
$x_{r}$ is typically generated from a user question $q$ by a search query generator $f$, such as a small language model.

\[
C = \mathcal{R}(x_{r}) \quad x_r = f(q)
\]
  
$\mathcal{G}$ produces an answer $y$ based on the input context $x_{g}$. 
$x_g$ typically includes a task instruction $\tau$, a user query $q$, and the retrieved chunks $C$.
\[
y = \mathcal{G}(x_{g}), \quad \text{where } x_g=(\tau, q, C)
\]

$C$ is implicitly divided into a relevant chunk set $C^{+}$, which contains evidence for generating an answer to $q$, and an irrelevant chunk set $C^{-}$ that does not contain supporting evidence.
From the perspective of retrieval performance, $\mathcal{R}$ does not guarantee retrieving all relevant chunks in $C^{+}$~\citep{MTEB}.
Therefore, $\mathcal{G}$ needs to appropriately extract evidence relevant to $q$ from $C^{+}$ within $C$ and generate an answer.

%%%%%%%%%%%%%%%%%%%%%%%%%%%%%%%%%%%%%%%%%%%%%%%%%%%%%%%%%%%%
%%%%%%%%%%%%%%%%%%%%%  Related Work  %%%%%%%%%%%%%%%%%%%%%%%
%%%%%%%%%%%%%%%%%%%%%%%%%%%%%%%%%%%%%%%%%%%%%%%%%%%%%%%%%%%%
\section{Related Work}
In recent years, various benchmarks have been proposed to systematically evaluate the performance of $\mathcal{G}$.
FRAMES~\citep{frames} provides an integrated, end-to-end evaluation of both $\mathcal{R}$ and $\mathcal{G}$ capabilities across three aspects—factuality, retrieval, and reasoning. The tasks require multi-document integration and involve temporal and numerical reasoning. This framework highlights how factual consistency and reasoning depth can be jointly measured under controlled retrieval conditions.
RAGBench~\citep{friel2025ragbenchexplainablebenchmarkretrievalaugmented} proposes the TRACe framework, which assesses $\mathcal{G}$ performance along three interpretable dimensions: Utilization (how much retrieved context is actually used), Adherence (faithfulness and hallucination control relative to the context), and Completeness (coverage of relevant information). It also measures retriever Relevance separately, enabling isolation of retrieval and generation errors.
RAGTruth~\citep{ragtruth} provides a dataset for analyzing and detecting hallucinations in RAG, defining four fine-grained types—Evident/Subtle Conflict and Evident/Subtle Baseless Information—to evaluate whether model outputs remain consistent with supporting documents. This benchmark highlights the difficulty of detecting subtle inconsistencies that remain semantically plausible but are factually incorrect.
RGB~\citep{rgb} evaluates $\mathcal{G}$ along four axes: noise robustness, abstention or negative rejection, multi-document information integration, and counterfactual robustness against misinformation. It tests models under varied, noisy, or conflicting evidence, providing insights into their ability to extract, integrate, or abstain appropriately when uncertainty arises.

Although these benchmarks contribute valuable insights, they primarily address limited aspects of $\mathcal{G}$’s behavior or evaluate each skill in isolation. 
In practical RAG applications, models must often interpret complex tables and perform multi-step reasoning simultaneously—for example, combining numerical computation with multi-hop inference across heterogeneous contexts.
Existing work has not yet captured this compound complexity, leaving a gap in evaluating $\mathcal{G}$’s real-world robustness.
To address this gap, our benchmark systematically assesses $\mathcal{G}$’s performance under such intertwined conditions, enabling more realistic and comprehensive evaluation of $\mathcal{G}$ capabilities required for practical RAG scenarios.

% \begin{table}
% \small
% \centering
% \caption{}
% \label{tab:related_benchmarks}
%     \begin{tabular}{|l|c|c|c|c|c|}
%     \hline
%         Dataset & Integration & Reasoning & Logic & Table & Abstention \\
%         \hline
%         \textbf{LIT-RAGBench} (our work) & \textcolor{green}{\checkmark} & \textcolor{green}{\checkmark} & \textcolor{green}{\checkmark} & \textcolor{green}{\checkmark} & \textcolor{green}{\checkmark} \\
%         \hline
%         TruthfulQA (Lin et al., 2021) & \textcolor{green}{\checkmark} & \textcolor{red}{\ding{55}} & \textcolor{red}{\ding{55}} & \textcolor{red}{\ding{55}} & \textcolor{red}{\ding{55}} \\
%         \hline
%     \end{tabular}
% \end{table}

%%%%%%%%%%%%%%%%%%%%%%%%%%%%%%%%%%%%%%%%%%%%%%%%%%%%%%%%%%
%%%%%%%%%%%%%%%%%%%%%  LIT-RAGBench  %%%%%%%%%%%%%%%%%%%%%%%
%%%%%%%%%%%%%%%%%%%%%%%%%%%%%%%%%%%%%%%%%%%%%%%%%%%%%%%%%%
\section{LIT-RAGBench}
%%%%%%%%%%%%%%%%%%%%%  Evaluation Framework  %%%%%%%%%%%%%%%%%%%%%%%
\subsection{Evaluation Framework}
\subsubsection{Evaluation Categories and Aspects}
LIT-RAGBench systematizes the core capabilities of $\mathcal{G}$ into five evaluation categories (\autoref{fig:proposal}), with each category subdivided into evaluation aspects based on practical use cases.
The five evaluation categories are: (1) \emph{Integration}, (2) \emph{Reasoning}, (3) \emph{Logic}, (4) \emph{Table}, and (5) \emph{Abstention}.
The first four categories, collectively referred to as \emph{Main}, represent the core capabilities required for $\mathcal{G}$ to generate a correct answer to $q$.
\emph{Abstention} is defined as an \emph{exceptional category}, distinct from \emph{Main}, as it evaluates $\mathcal{G}$’s ability to withhold an answer appropriately.

%% Integration
\paragraph{(1) Integration.}
This category addresses cases where evidence is dispersed across multiple documents, requiring $\mathcal{G}$ to extract and integrate relevant information from each source.
This capability has been examined in existing benchmarks, such as Integration in RGB~\citep{rgb} and Multiple Constraints in FRAMES~\citep{frames}. 
This category focuses on \emph{integrating information from multiple sources} ($|C^{+}|\geq 2$) as an evaluation aspect.
Single-source extraction ($|C^{+}|=1$) is a fundamental RAG operation that naturally co-occurs with other aspects and is not treated independently.
LIT-RAGBench targets integration from $2 \leq |C^{+}| \leq 3$ sources for simplicity and practicality.

%% Reasoning
\paragraph{(2) Reasoning.}
This category assesses $\mathcal{G}$'s reasoning capabilities across two dimensions.
% \emph{Multi-hop Reasoning} evaluates progressively, combining information from multiple documents to reach conclusions not explicitly stated in any single source, using benchmarks like HotpotQA~\citep{hotpotqa} and JEMHopQA~\citep{jemhopqa}.
\emph{Multi-hop Reasoning} evaluates $\mathcal{G}$'s ability to combine information from multiple documents to reach conclusions not explicitly stated in any single source. Questions were created using benchmarks such as HotpotQA~\citep{hotpotqa} and JEMHopQA~\citep{jemhopqa}.
\emph{Numerical Calculation} assesses deriving metrics (totals, averages, profit margins, growth rates) through common-sense arithmetic and business knowledge when formulas are not provided—a challenge for LLMs, compared to language tasks~\citep{math_1, math_2}.

%% Logic
\paragraph{(3) Logic.}
This category evaluates $\mathcal{G}$'s ability to interpret logical and linguistic relations between query $q$ and retrieved contexts $C$ despite lexical or semantic discrepancies.
Since $\mathcal{E}$ documents are typically unknown to users, phrasing mismatches frequently arise, requiring $\mathcal{G}$ to resolve them through logical understanding across three aspects.
%% Synonym Interpretation
\emph{Synonym Interpretation} recognizes equivalent expressions (e.g., "10 thousand yen" and "10,000 yen"), including multilingual terms or abbreviations, with LIT-RAGBench emphasizing numerical and unit expressions common in practical RAG scenarios.
%% Numerical Inclusion Interpretation
\emph{Numerical Inclusion Interpretation} assesses understanding of numerical conditions—e.g., determining whether a 35-year-old meets "20 or older and under 40" requires boundary-inclusive reasoning.
%% Conceptual Inclusion Interpretation
\emph{Conceptual Inclusion Interpretation} evaluates recognizing hierarchical relations, such as identifying "noise-canceling earphones" as "electronic devices" prohibited under device-banning rules.

% \begin{tcolorbox}[colback=green!5,colframe=green!25,coltitle=black!80,title=\textbf{Example (Logic)}]
%     \textbf{Q:} question question question question question question question question question question question question question? \\
%     \textbf{C1:} fugafuga1 fugafuga1 fugafuga1 fugafuga1 fugafuga1 fugafuga1 fugafuga1 fugafuga1 fugafuga1 fugafuga1 fugafuga1 fugafuga1 fugafuga1 fugafuga1 fugafuga1 fugafuga1 fugafuga1 fugafuga1 fugafuga1 fugafuga1 fugafuga1 fugafuga1 fugafuga1 fugafuga1 fugafuga1 ~ 
%     \textbf{C2:} fugafuga2 fugafuga2 fugafuga2 fugafuga2 fugafuga2 fugafuga2 fugafuga2 fugafuga2 fugafuga2 fugafuga2 fugafuga2 fugafuga2 fugafuga2 fugafuga2 fugafuga2 fugafuga2 fugafuga2 fugafuga2 fugafuga2 fugafuga2 fugafuga2 fugafuga2 fugafuga2 fugafuga2 fugafuga2 \\
%     \textbf{Expected:} Yes
% \end{tcolorbox}
% \vspace{-5mm}

%% Table
\paragraph{(4) Table.}
This category evaluates $\mathcal{G}$'s ability to interpret and extract information from tabular formats in retrieved contexts $C$.
RAG documents often contain tables in structured formats (HTML, Markdown, CSV) alongside text, requiring $\mathcal{G}$ to understand table structure and identify relevant data—capabilities assessed by benchmarks like MMQA~\citep{wu2025mmqa}.
\emph{HTML} tables use standard tags requiring $\mathcal{G}$ to map header-data relationships and comprehend hierarchical associations.
\emph{HTML with merged cells} employs \texttt{rowspan} or \texttt{colspan} attributes, creating multi-row/column dependencies that complicate positional relationships, challenging LLMs~\citep{Zhao_Cell-merged,Sui_Cell-merged}.
\emph{Markdown} tables use pipe delimiters with lightweight syntax but lack explicit typing or hierarchy, requiring $\mathcal{G}$ to infer row-column relationships without structural cues.
\emph{CSV} data provide comma/tab-separated values without headers or formatting, necessitating schema and semantic inference from contextual information alone.

%% Abstention
\paragraph{(5) Abstention.}
This category evaluates $\mathcal{G}$'s ability to refrain from answering when sufficient evidence is unavailable.
Withholding answers under inappropriate conditions is essential for response reliability and mitigating hallucinations~\citep{Hallucination_survey}, with benchmarks like AbstentionBench~\citep{abstentionbench} and RGB's negative rejection task~\citep{rgb}.
We define this capability across three aspects.
\emph{Insufficient Evidence} occurs when retrieved contexts lack necessary information; since $\mathcal{R}$ doesn't guarantee $C^{+}$, when $x_g = (\tau, q, C^{-})$, $\mathcal{G}$ should explicitly indicate it cannot answer, as LLMs tend to hallucinate without supporting evidence~\citep{rgb}.
\emph{Contradictory Evidence} arises when documents provide conflicting information, requiring $\mathcal{G}$ to recognize inconsistencies and either abstain or explicitly identify contradictions through cross-document consistency checking.
\emph{Incomplete Chunk} represents scenarios where retrieved content is fragmentary due to chunking boundaries splitting semantically connected information; when $\mathcal{R}$ retrieves partial segments—exceptionally long definitions or tables divided across boundaries—$\mathcal{G}$ should refrain from answering despite overlapping tokens potentially mitigating this issue.

\subsubsection{Category Composition}
This section formalizes the evaluation categories and aspects defined above.
Let $\Theta$ denote the set of five evaluation categories treated in this study. 
The abbreviations $\theta_I$, $\theta_R$, $\theta_L$, $\theta_T$, and $\theta_A$ are used for \emph{Integration}, \emph{Reasoning}, \emph{Logic}, \emph{Table}, and \emph{Abstention}, respectively. 
Collectively, these are expressed as $\Theta = {\theta_I, \theta_R, \theta_L, \theta_T, \theta_A}$.
We define $\Theta_{\text{Main}} = \Theta \setminus \{\theta_A\}$. 
$\theta_A$ is independent and does not co-occur with aspects from $\Theta_{\text{Main}}$.
Each evaluation category $\theta \in \Theta$ comprises multiple evaluation aspects associated with it.
Let $\Phi$ denote the set of evaluation aspects, and let $\Phi_{\theta}$ represent the subset of aspects belonging to a particular evaluation category $\theta$.
Then, $\Phi$ is defined as the union of all $\Phi_{\theta}$ for $\theta \in \Theta$.

LIT-RAGBench is constructed by combining evaluation aspects belonging to one or two categories. 
For each evaluation problem $q$, the set of aspects $\psi(q)$ satisfies:
\begin{equation*}
  \Psi(q) \subseteq \Phi,\quad 1 \leq |\Psi(q)| \leq 2
\end{equation*}
\begin{equation*}
  \forall \phi_i, \phi_j \in \Psi(q),\quad \phi_i \in \Phi_{\theta_m}, \phi_j \in \Phi_{\theta_n} \Rightarrow m \neq n
\end{equation*}

Thus, category composition is formalized through the aspect $\Psi(q)$ and the non-overlap constraint across categories.
An example of co-occurrence that satisfies these constraints is shown in ~\autoref{fig:cooccur-example}.

Consequently, LIT-RAGBench enables comprehensive and realistic evaluation of the multifaceted capabilities of $\mathcal{G}$, as well as quantitative analyses across categories and aspects.

\begin{figure}[t]
    \centering
    \small
    \begin{tcolorbox}[
      title={\textbf{\emph{Reasoning (Multi-hop)} \\ $\times$ \\ \emph{Table (HTML with merged cells)}}},
      width=\linewidth,
      fontupper=\scriptsize,
      left=2pt,
      right=2pt,
      top=2pt,
      bottom=2pt
    ]
        \textbf{$q$}: What major research theme was officially adopted by \underline{GreenWave} in 2024 regarding the marine plastic issue?
        
        \vspace{3pt}
        \textbf{$C^{+}_{1}$}: 
        GreenWave concluded a partnership with \underline{Seikai University}; in Jul 2024, the joint project was selected for the Ministry of the Environment's \underline{Blue Innovation Research Grant}. $\cdots$
        
        \vspace{3pt}
        \textbf{$C^{+}_{2}$}: List of Blue Innovation Research Grant 
        
        \begin{tabular}{@{}p{0.12\linewidth}p{0.22\linewidth}p{0.52\linewidth}@{}}
        \toprule
        Univ. & Field & Summary \\
        \midrule
        \multirow{2}{=}{Seikai Univ.} 
          & Marine plastics
          & \textbf{Development of biopolymers that decompose microplastics} \\
        \cmidrule{2-3}
          & Marine ecosystems 
          & Assessment of impact of drifting debris on coastal organisms \\
        \midrule
        Toyo Inst. 
          & Hydrogen energy 
          & Optimization of algae-derived hydrogen production system \\
        \bottomrule
        \end{tabular}
        
        \vspace{3pt}
        
        \textit{Rationale}: From $C^{+}_{1}$, the GreenWave $\leftrightarrow$ Seikai Univ.\ project was adopted; the merged-cell row in $C^{+}_{2}$ gives the adopted theme.
        
        \textbf{$a$}: Development of a biopolymer that degrades microplastics
    \end{tcolorbox}
    \caption{Example of a co-occurrence across evaluation categories ($\theta_R$ and $\theta_T$)}
  \label{fig:cooccur-example}
\end{figure}

%%%%%%%%%%%%%%%%%%%%%  Dataset Construction  %%%%%%%%%%%%%%%%%%%%%%%
\subsection{Dataset Construction} 
The dataset $D$ consists of $m$ samples, each containing a question $q_i$, its answer $a_i$, and document sets $C_i^{+}$ and $C_i^{-}$.
\begin{equation*}
  D = \{(q_i, a_i, C_i^{+}, C_i^{-}, \Psi(q_i)) \mid 1 \leq i \leq m\}.
\end{equation*}
% Here, $C_i^{+}$ includes supporting evidence, while $C_i^{-}$ contains irrelevant chunks.
% Both sets are either manually created or mechanically synthesized texts that imitate the output of $\mathcal{R}$ and have passed manual inspection.
% To ensure that each input $x_g$ provides sufficient contextual information, we impose the constraint $|C_i^{+}| + |C_i^{-}| \geq 8$.
To ensure sufficient context, $|C_i^{+}| + |C_i^{-}| \geq 8$ is required.

During evaluation, the task instruction $\tau$ specifies two directives: (1) answer based on the given context, and (2) state inability to answer when supporting evidence is unavailable.
For $\Theta_{\text{Main}}$, answers are generated with $x_g = (\tau, q_i, C_i)$, where $C_i = C_i^{+} \cup C_i^{-}$.
We randomize the concatenation order of $C_i$ at input time. This removes confounding from retriever ranking, mitigates position bias~\citep{lost_in_the_middle}, and evaluates the use of order-invariant evidence and robustness to rank perturbations across retrievers.
% In the evaluation of the \emph{Insufficient Evidence} aspect in $\theta_A$, answers are generated using $x_g = (\tau, q_i, C_i^{-})$, which excludes any descriptions related to evidence. 
% In this case as well, the order of $C_i^{-}$ is randomly shuffled.
For the \emph{Insufficient Evidence} evaluation in $\theta_A$, we use $x_g = (\tau, q_i, C_i^{-})$, again with randomized order.

% 作問手順
\subsubsection{Dataset Creation Procedure}
\label{sec:DatasetCreationProcedure}
The dataset was manually created by three native Japanese speakers. They designed question–answer scenarios and constructed relevant document sets following predefined guidelines. For quality assurance, two of the three contributors independently reviewed all items, and only samples that passed both inspections were retained in the final dataset.
To enhance the efficiency of the dataset creation process, we additionally used GPT-5 as an auxiliary tool (see \footnoteref{footnote_repo} for the prompt templates used).

% QA scenariosの決定
\paragraph{Decision QA scenarios.}
QA scenarios were designed based on evaluation aspect patterns, assuming practical RAG use cases. Drawing on the methodology of Kirchenbauer et al.~\citep{kirchenbauer2025fictionalqadatasetstudying}, QA scenarios were created using fictional knowledge entities like company names, product names, and personal names to prevent LLMs from answering based on their pre-trained knowledge when generating answers. 

% 質問文 $q_i$・関連文書集合 $C_i^{+}$・正答文章 $a_i$の作成
\paragraph{Creation of questions, relevant document sets, and answers.}
% In this process, we used OpenAI's GPT-5 to enhance the efficiency of question creation in QA scenarios. 
The automatically generated texts were manually reviewed for quality, and only those judged to appropriately reflect the evaluation aspects were adopted. 
Additionally, the length of each document in $C_i^{+}$ was adjusted to approximately 512 tokens using tiktoken \footnote{\url{https://github.com/openai/tiktoken}} to align with typical chunk lengths.

% 非関連文書集合 $C_i^{-}$の作成
\paragraph{Creation of irrelevant document sets.}
Assuming the output of $\mathcal{R}$, $C^{-}$ are designed to be related to $q_i$ or $C_i^{+}$, for example, by containing keywords from $q_i$, even though they do not serve as direct evidence. Following the same procedure as $C^{+}$, manual quality inspection and token length adjustment were conducted.

% 人手によるフィルタリング
\paragraph{Human-based filtering.}
For the obtained $q_i$, $a_i$, $C_i^{+}$, $C_i^{-}$, and $\Psi(q_i)$, verification was conducted from the following perspectives: (1) whether the question appropriately corresponds to the target evaluation aspect pattern, (2) whether it is possible to derive $a_i$ from $C_i^{+}$ as evidence for $q_i$, (3) whether the problem can not be answered using only the LLM's pre-trained knowledge, and (4) whether the fictional information contradicts real facts. 
Only problems that satisfied all criteria were added to $D$. In cases of non-compliance, the problem was discarded, and the procedure was repeated from scenario design in step 1. 
% The manual verification was conducted by the authors, all of whom are native Japanese speakers. （カメラレディではこっちにする）
The manual verification was conducted by qualified annotators who are native Japanese speakers.

\subsubsection{Statistics}
Following the creation procedure described above, a 54-question Japanese QA dataset was constructed and included in $\Theta_{\text{Main}}$. 
There are 12 questions with $|\Psi(q_i)|=1$ and 42 questions with $|\Psi(q_i)|=2$. 
From this 54-question Japanese QA dataset, an additional 54 QA questions for the \emph{Insufficient Evidence} aspect of $\theta_{A}$ were created by converting $C^{+}$ to an empty set. 
For the \emph{Contradictory Evidence} and \emph{Incomplete Chunk} aspects of $\theta_A$, three questions were randomly selected from the constructed $\Theta_{\text{Main}}$ QA dataset for each aspect, and $C^{+}$ was manually edited to construct the corresponding QA data. 
This process resulted in a Japanese evaluation dataset with a total of 114 questions. 
An English evaluation dataset of the same scale was constructed by translating the Japanese dataset using the GPT-5 API-based LLM.
% \autoref{tab:evaluation_categories} shows the number of questions for each evaluation aspect. 
% Since some questions have compound aspects ($|\psi(q)|=2$), the aspect counts are not mutually exclusive, and their sum exceeds the total number of questions.

For $\theta_{A}$, the aspect distribution was \emph{Insufficient Evidence} 34.6\%, \emph{Contradictory Evidence} 1.9\%, and \emph{Incomplete Chunk} 1.9\%. 
The proportion of \emph{Insufficient Evidence} is relatively large because these instances can be directly created by removing $C^{+}$ from the corresponding $\Theta_{\text{Main}}$.

% \begin{table}[t]
%     \centering
%     \caption{Percentage of Questions by Evaluation Aspect in LIT-RAGBench}
%     \small
%     \resizebox{\columnwidth}{!}{%
%     \begin{tabular}{l l r}
%         \toprule
%         \textbf{Category} & \textbf{Aspect} & \textbf{\%} \\ 
%         \midrule
%         Integration & - & 7.7 \\
%         \midrule
%         \multirow{2}{*}{Reasoning}
%         & Multi-hop Reasoning & 7.7 \\
%         & Numerical Calculation & 7.1 \\
%         \midrule
%         \multirow{3}{*}{Logic}
%         & Synonym Interp. & 7.1 \\ 
%         & Num. Inclusion Interp. & 6.4 \\ 
%         & Concept. Inclusion Interp. & 5.8 \\ 
%         \midrule
%         \multirow{4}{*}{Table}
%         & HTML & 5.1 \\
%         & HTML w/ merged cells & 5.8 \\ 
%         & Markdown & 4.5 \\
%         & CSV & 4.5 \\
%         \midrule
%         \multirow{3}{*}{Abstention}
%         & Insufficient Evidence & 34.6 \\
%         & Contradictory Evidence & 1.9 \\
%         & Incomplete Chunk & 1.9 \\
%         \bottomrule
%     \end{tabular}
%     }
%     \label{tab:evaluation_categories}
% \end{table}

%%%%%%%%%%%%%%%%%%%%%%%%%%%%%%%%%%%%%%%%%%%%%%%%%%%%%%%%%
%%%%%%%%%%%%%%%%%%%%%  Experiments  %%%%%%%%%%%%%%%%%%%%%
%%%%%%%%%%%%%%%%%%%%%%%%%%%%%%%%%%%%%%%%%%%%%%%%%%%%%%%%%
\section{Experiments}
In this section, we present the evaluation results and analysis of LLMs using LIT-RAGBench. We evaluate both API-based and open-weight models on the Japanese and English datasets of LIT-RAGBench.

%%%%%%%%%%%%%%%%%%%%%  Experimental Settings  %%%%%%%%%%%%%%%%%%%%%
\subsection{Experimental Settings}
\autoref{table:evaluated_models} lists the evaluated API-based and open-weight models, respectively.
For Reasoning Language Models (RLMs) such as GPT-5 and o3, the length of reasoning token generation can be specified, and in all cases, the maximum generation length setting is used.
For models that support temperature configuration, \texttt{temperature} is set to 0.0 and \texttt{top\_p} to 1.0.

\begin{table}[t]
  \centering
  \small
  \caption{Evaluated LLMs (API-based and open-weight, $^{\dagger}$ denotes reasoning models)}
  \resizebox{\columnwidth}{!}{%
  \begin{tabular}{l l l}
    \toprule
    \textbf{Model Name} & \textbf{Version} & \textbf{Developer} \\
    \midrule
    \multicolumn{3}{l}{\emph{API Models}} \\
    \midrule
    GPT-5$^{\dagger}$ & 2025-08-07 & OpenAI \\
    GPT-5-mini$^{\dagger}$ & 2025-08-07 & OpenAI \\
    GPT-5-nano$^{\dagger}$ & 2025-08-07 & OpenAI \\
    o3$^{\dagger}$ & 2025-04-16 & OpenAI \\
    o4-mini$^{\dagger}$ & 2025-04-16 & OpenAI \\
    GPT-4.1 & 2025-04-14 & OpenAI \\
    GPT-4.1-mini & 2025-04-14 & OpenAI \\
    Gemini-2.5-Flash & 2025-05-17 & Google \\
    Gemini-2.5-Pro$^{\dagger}$ & 2025-06-17 & Google \\
    Claude-Sonnet-4 & 2025-05-14 & Anthropic \\
    \midrule
    \multicolumn{3}{l}{\emph{Open Models}} \\
    \midrule
    Gemma-3-27B-Instruct\footnotemark[2] & -- & Google \\
    Llama-3.1-8B-Instruct\footnotemark[3] & -- & Meta \\
    Llama-3.3-70B-Instruct\footnotemark[4] & -- & Meta \\
    Qwen3-235B-A22B-Instruct\footnotemark[5] & -- & Alibaba \\
    Qwen3-235B-A22B-Thinking$^{\dagger}$\footnotemark[6] & -- & Alibaba \\
    \bottomrule
  \end{tabular}%
  }
  \label{table:evaluated_models}
\end{table}

\footnotetext[2]{\texttt{google/gemma-3-27b-it}}
\footnotetext[3]{\texttt{meta-llama/Llama-3.1-8B-Instruct}}
\footnotetext[4]{\texttt{meta-llama/Llama-3.3-70B-Instruct}}
\footnotetext[5]{\texttt{Qwen/Qwen3-235B-A22B-Instruct-2507}}
\footnotetext[6]{\texttt{Qwen/Qwen3-235B-A22B-Thinking-2507}}

%%% Evaluation Method
\subsection{Evaluation Method}
% To evaluate the accuracy of answers generated by $\mathcal{G}$, this study employs automated evaluation using an LLM-as-a-Judge.
% Previous studies have demonstrated a strong correlation between LLM-as-a-Judge evaluations and human judgments~\citep{llm_as_a_judge,g-eval,prometheus}.
% Let $\mathcal{J}$ denote the LLM-based evaluator.
% Given the inputs $q$, $a$, and $y$, $\mathcal{J}$ determines the correctness of $a$ based on its consistency with $y$. As a result, it outputs a binary label: ``1'' if correct and ``0'' if incorrect.
% In this study, $\mathcal{J}$ is implemented using GPT-4.1.
% \begin{equation*}
%   \mathcal{J}(q_i, a_i, y_i) = \begin{cases}
%     1 & \text{if } y_i = a_i \\
%     0 & \text{otherwise}
%   \end{cases}
% \end{equation*}

To evaluate the accuracy of answers generated by $\mathcal{G}$, this study employs automated evaluation using an LLM-as-a-Judge. 
The prompt used for the LLM-as-a-Judge is provided in the supplementary repository (see~\footnoteref{footnote_repo}).
While LLM-based automatic evaluation methods have raised concerns about reliability, our evaluation task is a relatively straightforward closed-ended comparison between the generated answer and the reference answer, which is more amenable to automated judgment.
Previous studies have demonstrated a strong correlation between LLM-as-a-Judge evaluations and human judgments~\citep{llm_as_a_judge,g-eval,prometheus}.
% To evaluate the accuracy of answers generated by $\mathcal{G}$, this study employs automated evaluation using an LLM-as-a-Judge.
% Previous studies have demonstrated a strong correlation between LLM-as-a-Judge evaluations and human judgments~\citep{llm_as_a_judge,g-eval,prometheus}.
Let $\mathcal{J}$ denote the LLM-based evaluator.
In this study, we implement $\mathcal{J}$ using GPT-4.1 (OpenAI API, version dated 2025-04-14), which is prompted to determine whether the generated answer $y_i$ is semantically consistent with the reference answer $a_i$.
$\mathcal{J}$ outputs a binary label for each instance:
\begin{equation*}
  \mathcal{J}(q_i, a_i, y_i) = 
  \begin{cases}
    1 & \text{if } y_i \text{ is consistent with } a_i \\
    0 & \text{otherwise}
  \end{cases}
\end{equation*}

This study uses accuracy as the performance metric. 
Let $Q$ and $Q_{\theta}$ denote the set of all questions and the subset belonging to category $\theta \in \Theta$, respectively.
Accuracy for each category is defined as:
\begin{equation*}
  \text{Accuracy}(\theta) = \frac{1}{|Q_{\theta}|}\sum_{q_i \in Q_{\theta}} \mathcal{J}(q_i, a_i, y_i)
\end{equation*}
The overall accuracy, denoted as $\overline{\text{Accuracy}}$, is obtained by averaging $\text{Accuracy}(\theta)$ over all $\theta$.
When a question involves multiple aspects and is answered correctly, it is counted as correct for all applicable aspects.

%%% 4.2 評価結果
\subsection{Results}
\begin{figure*}[t]
  \centering
  \includegraphics[width=\textwidth]{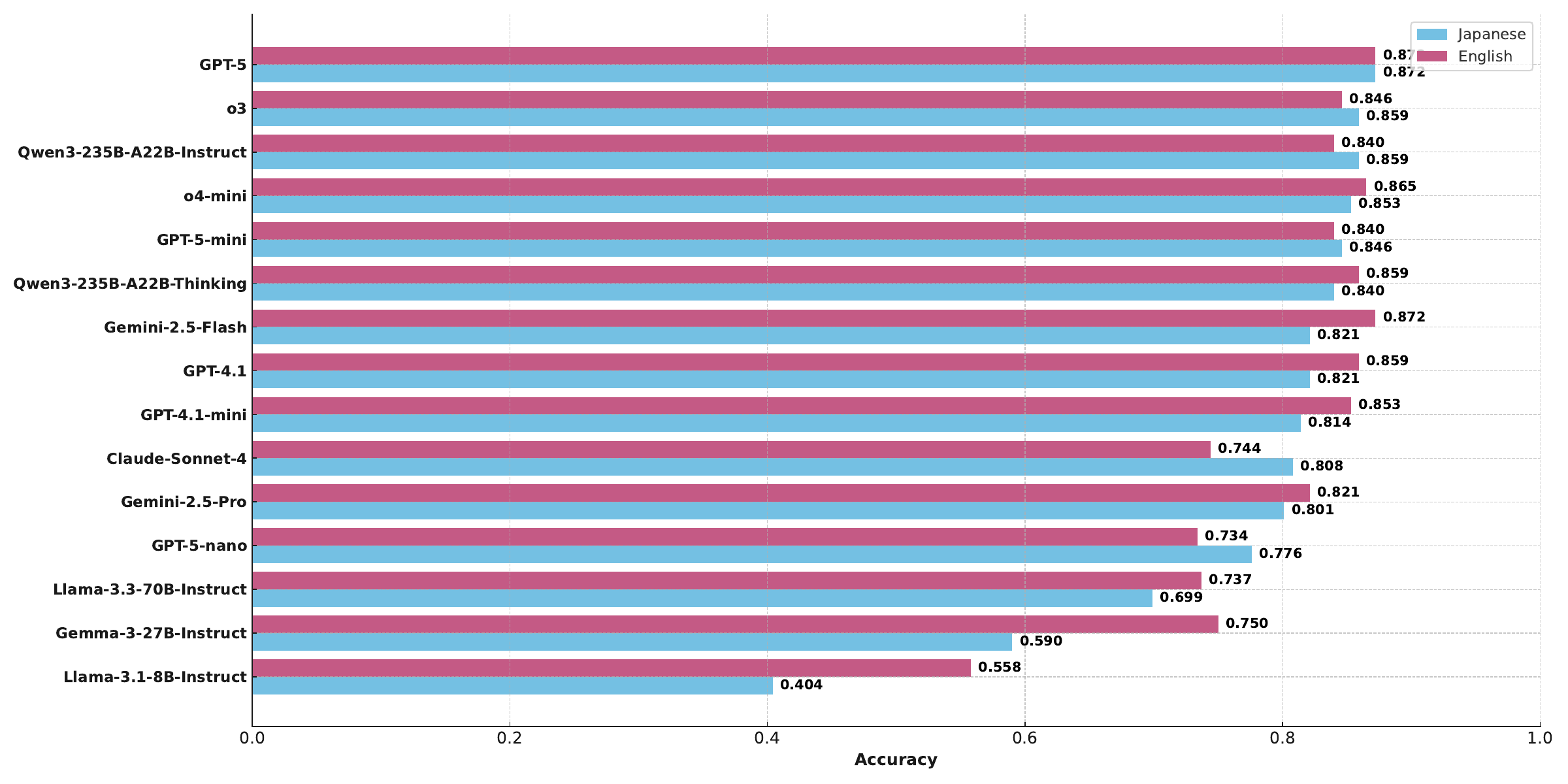}
  \caption{Evaluation results for overall accuracy. Blue bars represent Japanese scores, and red bars represent English scores.}
  \label{fig:accuracy_ja_en}
\end{figure*}

The evaluation results for $\overline{\text{Accuracy}}$ on the Japanese and English datasets are shown in \autoref{fig:accuracy_ja_en}. 
\autoref{table:accuracy_categories} presents $\text{Accuracy}(\theta)$ for each $\theta \in \Theta$.

\begin{table*}[t]
    \centering
    \footnotesize
    \setlength{\tabcolsep}{3pt}
    \caption{Category-wise accuracy comparison across models for Japanese and English evaluations. The best and second-best performances in each column are highlighted in bold and underlined, respectively. In cases of ties, all models achieving the same best or second-best score are highlighted accordingly.}
    \resizebox{\textwidth}{!}{%
    \begin{tabular}{l|cc|cc|cc|cc|cc|cc}
        \toprule
        \multirow{2}{*}{Model} &
        \multicolumn{2}{c|}{\cellcolor{red!20}\textbf{Integration}} &
        \multicolumn{2}{c|}{\cellcolor{yellow!25}\textbf{Reasoning}} &
        \multicolumn{2}{c|}{\cellcolor{green!15}\textbf{Logic}} &
        \multicolumn{2}{c|}{\cellcolor{purple!25}\textbf{Table}} &
        \multicolumn{2}{c|}{\cellcolor{gray!40}\textbf{Main (avg)}} &
        \multicolumn{2}{c}{\cellcolor{blue!15}\textbf{Abstention}} \\
        & ja & en & ja & en & ja & en & ja & en & ja & en & ja & en \\
        \midrule
        \multicolumn{13}{l}{\emph{API Models}} \\
        \midrule
        GPT-5 & \underline{0.833} & 0.833 & 0.870 & 0.739 & \underline{0.867} & \textbf{0.900} & \underline{0.839} & 0.839& 0.852 & 0.828& \underline{0.900} & \underline{0.933} \\
        GPT-5-mini & \textbf{0.917} & 0.750 & 0.826 & 0.696 & \underline{0.867} & \underline{0.867} & 0.774 & 0.806 & 0.846 & 0.780 & 0.883 & 0.917 \\
        GPT-5-nano & 0.750 & 0.667 & 0.565 & 0.739 & 0.733 & 0.833 & 0.677 & 0.710 & 0.681 & 0.737 & 0.767 & 0.817 \\
        o3 & \underline{0.833} & 0.833 & \textbf{0.957} & \textbf{0.870} & \textbf{0.900} & \textbf{0.900} & \underline{0.839} & 0.806 & \underline{0.882} & \underline{0.852} & 0.817 & 0.833 \\
        o4-mini & \textbf{0.917} & 0.833 & \underline{0.913} & 0.783 & \textbf{0.900} & \textbf{0.900} & \textbf{0.871} & 0.839& \textbf{0.900} & 0.839 & 0.783 & 0.900 \\
        GPT4.1 & \underline{0.833}& \textbf{1.000} & 0.739 & 0.739 & 0.800 & 0.800 & \underline{0.839} & 0.806 & 0.803 & 0.836 & 0.850 & \underline{0.933} \\
        GPT4.1 mini & \textbf{0.917} & 0.833 & 0.870 & \underline{0.826} & 0.800 & 0.833 & 0.806 & \underline{0.871} & 0.848 & 0.841& 0.783 & 0.867 \\
        Claude-Sonnet-4 & 0.750 & \underline{0.917} & 0.783 & 0.565 & 0.700 & 0.600 & 0.677 & 0.516 & 0.728 & 0.650 & \textbf{0.950} & \textbf{0.967} \\
        Gemini-2.5-Flash & \textbf{0.917} & \underline{0.917} & 0.783 & \underline{0.826} & \underline{0.867}& 0.833 & \textbf{0.871} & \textbf{0.903} & 0.860 & \textbf{0.870} & 0.767 & 0.883 \\
        Gemini-2.5-Pro & 0.667 & 0.833 & 0.870& 0.696 & 0.833 & 0.833 & 0.774 & 0.774 & 0.786 & 0.784 & 0.800 & 0.883 \\
        \midrule
        \multicolumn{13}{l}{\emph{Open Models}} \\
        \midrule
        Gemma-3-27B-Instruct & 0.667 & 0.667 & 0.348 & 0.609 & 0.567 & 0.733 & 0.484 & 0.710 & 0.517 & 0.680 & 0.733 & 0.850 \\
        Llama-3.1-8B-Instruct & 0.167 & 0.500 & 0.130 & 0.261 & 0.367 & 0.500 & 0.355 & 0.323 & 0.255 & 0.396 & 0.600 & 0.833 \\
        Llama-3.3-70B-Instruct & 0.750 & 0.583 & 0.478 & 0.609 & 0.733 & 0.700 & 0.677 & 0.581 & 0.660 & 0.618 & 0.767 & 0.917 \\
        Qwen3-235B-A22B-Instruct & \textbf{0.917} & 0.833 & 0.870 & 0.783 & 0.833 & \underline{0.867} & \underline{0.839} & 0.742 & 0.865 & 0.806 & 0.867 & 0.900 \\
        Qwen3-235B-A22B-Thinking & \textbf{0.917} & 0.750 & 0.826 & \textbf{0.870} & 0.767 & \underline{0.867} & 0.774 & 0.774 & 0.821 & 0.815& \underline{0.900} & 0.917 \\
        \bottomrule
    \end{tabular}
    }
    \label{table:accuracy_categories}
\end{table*}

As shown in \autoref{fig:accuracy_ja_en}, no model achieved an $\overline{\text{Accuracy}}$ of 0.9 in either the Japanese or English evaluations.
GPT-5 achieved the highest score of 0.872 in both languages. 
Among open-weight models, Qwen3-235B-A22B-Instruct and Qwen3-235B-A22B-Thinking showed the best performance, with scores of 0.859 and 0.821, respectively.
In contrast, open-weight models with small to medium parameter sizes, such as Llama-3.1-8B-Instruct and Gemma-3-27B-Instruct, showed generally low scores.

%%% 4.3 分析
\subsection{Analysis}

% \begin{figure*}[t]
%   \centering
%   \begin{subfigure}[t]{0.19\textwidth}
%     \includegraphics[width=\textwidth]{figures/LIT-RAGBench.png}
%     \caption{Integration}
%   \end{subfigure}
%   \hfill
%   \begin{subfigure}[t]{0.19\textwidth}
%     \includegraphics[width=\textwidth]{figures/LIT-RAGBench.png}
%     \caption{Reasoning}
%   \end{subfigure}
%   \hfill
%   \begin{subfigure}[t]{0.19\textwidth}
%     \includegraphics[width=\textwidth]{figures/LIT-RAGBench.png}
%     \caption{Logic}
%   \end{subfigure}
%   \hfill
%   \begin{subfigure}[t]{0.19\textwidth}
%     \includegraphics[width=\textwidth]{figures/LIT-RAGBench.png}
%     \caption{Table}
%   \end{subfigure}
%   \hfill
%   \begin{subfigure}[t]{0.19\textwidth}
%     \includegraphics[width=\textwidth]{figures/LIT-RAGBench.png}
%     \caption{Abstention}
%   \end{subfigure}
%   \caption{
%     Examples from each error type:
%     (a) Integration,
%     (b) Reasoning,
%     (c) Logic,
%     (d) Table, and
%     (e) Abstention.
%   }
% \end{figure*}

%%% Integration
\subsubsection{Integration}
% \autoref{table:accuracy_categories} shows that, in the $\theta_{I}$, many models achieved relatively high scores. Among them, Gemini-2.5-Flash, GPT-5-mini, o4-mini, GPT-4.1 mini, Qwen3-235B-A22B-Instruct, and Qwen3-235B-A22B-Thinking attained the highest performance of 0.917. Large language models such as GPT-5, o3, and GPT-4.1 followed with an accuracy of 0.833. This result indicates that smaller models, including GPT-5-mini and GPT-4.1 mini, outperformed larger ones such as GPT-5 and o3, despite the latter exhibiting strong performance on $\Theta_\text{Main}$.
When examining outputs that produced incorrect answers, errors frequently occurred when $C^{+}$ lacked explicit lexical cues or contained multiple similar pieces of information differing only in minor details. For example, when extracting prices from rate tables of three meeting room reservation systems—Company A, B, and C—only Company A included additional notes on pricing. Correct extraction required accounting for these notes, but the model returned only the basic fee, as with Companies B and C.
These observations indicate that $\mathcal{G}$ often fails when data sources vary in quality or granularity, reaffirming the importance of preprocessing, such as document structuring and normalization, in practical RAG deployments.

%%% Reasoning
\subsubsection{Reasoning}
% \autoref{table:accuracy_categories} shows that, in the $\theta_{R}$ category, o3 achieved the highest score of 0.957, correctly answering nearly all problems. Additionally, reasoning models, such as o4-mini, GPT-5, and Gemini-2.5-Pro, demonstrated higher scores compared to other models.
In the numerical calculation aspect, o3 solved all tasks correctly.
In contrast, o4-mini and GPT-5 generally followed appropriate reasoning steps but made minor arithmetic errors in intermediate or final calculations, consistent with previous findings~\cite{math_1, math_2}.
Small-to-medium-scale open-weight and API-based mini models showed limited reasoning capability. These models often failed to identify intermediate entities in the reasoning chain or refused to answer when explicit lexical cues were absent.

For instance, in a question requiring inference of a company's 2024 ranking from its 2025 position and relative improvement, such models recognized both facts but failed to infer the implicit 2024 value, resulting in incorrect responses.
This limitation frequently co-occurred with $\theta_T$, suggesting difficulty in integrating implicit relationships or reasoning across multiple documents rather than from explicitly stated facts, consistent with prior work~\citep{rag_multihop_problem}.

%%% Logic
\subsubsection{Logic}
As shown in \autoref{table:accuracy_categories}, most API-based models achieved high accuracy in both Japanese and English for $\theta_L$.
However, some category-specific errors were observed.
For instance, when asked about data capacity in GB, models answered “500 MB” from the source instead of the correct “0.5 GB.” Although the relevant text was correctly identified, such responses were judged incorrect under the benchmark criteria. These issues can be mitigated through prompts that enforce output-unit alignment, emphasizing the importance of operational rule design.
A Japanese-specific error was also found in unit conversion tasks common in business contexts: models often generated “760 million yen” instead of the correct “7.6 billion yen.” Addressing these language-dependent hallucinations is practically important, and future work will extend the benchmark to include such cases.

%%% Table
\subsubsection{Table}
From \autoref{table:accuracy_categories}, Gemini-2.5-Flash achieved the highest $\theta_T$ score in both Japanese and English. Most LLMs understood basic row–column structures but struggled with merged-cell tables, making accurate information retrieval difficult, especially for small-to-medium-scale open-weight models.

When large tables exceeding 512 tokens were split into chunks in $C^{+}$, nearly all models failed to extract relevant data. We anticipated this issue during dataset construction and inserted header information into each chunk during preprocessing. However, because the chunks were intentionally shuffled to simulate real-world disorder, models likely failed to recognize the overall structure and often abstained despite sufficient evidence.

These findings highlight that in practical applications, preprocessing such as table restructuring and reordering chunks is essential before inputting to $\mathcal{G}$, since large tables are often unintentionally split.

%%% Abstention
\subsubsection{Abstention}
From \autoref{table:accuracy_categories}, Claude-Sonnet-4 achieved notably high $\theta_A$ scores in both Japanese and English.
In the Insufficient Evidence category, numerous models produced knowledge that appeared contextually plausible but lacked evidential grounding. 
Such behavior has been identified in a previous study as a type of hallucination arising from fabricated knowledge~\citep{Hallucination_survey, abstentionbench}.
For Contradictory Evidence, Claude-Sonnet-4 detected inconsistencies in all tasks and correctly abstained, while many models relied on a single source and failed to cross-reference evidence.
In the Incomplete Chunk aspect, most models appropriately abstained, but when faced with ambiguous questions that could be answered using general knowledge, they often hallucinated responses.
Such ambiguity frequently arises in practical RAG scenarios, as users expect responses grounded in retrieved evidence $\mathcal{E}$.
These results underscore the importance of prompt design to guide appropriate abstention behavior in RAG systems.

Based on quantitative and qualitative analyses, Claude-Sonnet-4 demonstrates a strong ability to refrain from answering. 
However, this also reveals a tendency toward Over-Abstention even when a valid response is possible.
In this study, instances within $\theta_{\text{Main}}$ where models abstained despite having sufficient information—resulting in incorrect outcomes—were quantified as the Over-Abstention Rate (\autoref{table:over_abstention}).
Based on the average Over-Abstention Rate (\textit{avg} in \autoref{table:over_abstention}), notable disparities are observed across models.
Claude-Sonnet-4 shows the highest rate (0.259), followed by Llama-3.1-8B-Instruct (0.213), indicating a strong tendency to over-abstain.
% In contrast, GPT-5 and Gemini-2.5-Flash exhibit substantially lower averages (both 0.065), suggesting a more balanced approach between caution and responsiveness.
Among open-weight models, larger models such as Qwen3-235B-A22B-Thinking (0.120) maintain moderate rates, whereas smaller models like Llama-3.1-8B-Instruct tend to over-abstain, likely due to weaker reasoning capacity in $\Theta_{\text{Main}}$.
These patterns suggest that underdeveloped generation capabilities may lead models to default to abstention when uncertain, even in answerable cases.
In a previous study designed to measure over-abstention tendencies, a strong correlation between safety alignment and usefulness was also observed, indicating a clear trade-off between the two dimensions \citep{or_bench}. Consistent with our results, the study reported that Claude exhibited the strongest tendency toward over-abstention among the evaluated models. 
These results indicate that while a higher abstention rate may reflect cautious alignment behavior, it does not necessarily translate into greater overall usefulness.

\begin{table}[t]
\centering
\small
\setlength{\tabcolsep}{6pt}
\caption{Over-Abstention Rates on the Japanese and English Evaluation Sets for the Main Category. 
The column \textit{avg} indicates the mean of the Japanese (\textit{ja}) and English (\textit{en}) rates.}
\resizebox{\columnwidth}{!}{
    \begin{tabular}{l|ccc}
    \toprule
    Model & ja & en & avg \\
    \midrule
    Claude-Sonnet-4 & \underline{0.148} & \textbf{0.370} & \textbf{0.259} \\
    Llama-3.1-8B-Instruct & \textbf{0.204} & \underline{0.222} & \underline{0.213} \\
    Llama-3.3-70B-Instruct & 0.093 & \underline{0.222} & 0.157 \\
    Gemma-3-27B-Instruct & 0.167 & 0.074 & 0.120 \\
    Qwen3-235B-A22B-Thinking & 0.130 & 0.111 & 0.120 \\
    GPT-4.1 & 0.056 & 0.111 & 0.083 \\
    Qwen3-235B-A22B-Instruct & 0.056 & 0.111 & 0.083 \\
    GPT-5-mini & 0.074 & 0.074 & 0.074 \\
    GPT-5 & 0.037 & 0.093 & 0.065 \\
    Gemini-2.5-Flash & 0.074 & 0.056 & 0.065 \\
    GPT-5-nano & 0.074 & 0.019 & 0.046 \\
    o3 & 0.019 & 0.074 & 0.046 \\
    o4-mini & 0.019 & 0.074 & 0.046 \\
    Gemini-2.5-Pro & 0.000 & 0.093 & 0.046 \\
    GPT-4.1 mini & 0.037 & 0.019 & 0.028 \\
    \bottomrule
    \end{tabular}
}
\label{table:over_abstention}
\vspace{-3mm}
\end{table}

%%%%%%%%%%%%%%%%%%%%%  考察  %%%%%%%%%%%%%%%%%%%%% 
\subsection{Overall Discussion}
Experiments using LIT-RAGBench demonstrated that composite-category evaluations can quantitatively assess the multifaceted abilities of $\mathcal{G}$.
Unlike existing benchmarks that assess each category independently, LIT-RAGBench enables absolute comparisons across categories, revealing integrated performance differences and weaknesses among models.
The experiments also showed that LIT-RAGBench effectively captures variations in abstention tendencies.
While improved abstention behavior enhances response reliability, it may reduce accuracy on answerable questions.
These findings indicate that LLMs can benefit from further training and prompt optimization to balance abstention and accuracy.

% \begin{itemize}[leftmargin=0.5cm,topsep=0pt,itemsep=0pt]
%     \item \textbf{Composite-category evaluation:} 
%     Experiments using LIT-RAGBench confirmed that evaluation with composite-category questions can quantitatively assess the multifaceted abilities of $\mathcal{G}$.
    
%     \item \textbf{Absolute comparison across capabilities:}
%     While conventional benchmarks that evaluate each category independently enable relative comparisons of capabilities, LIT-RAGBench facilitates \textit{absolute comparisons across categories} through composite-category evaluations, thereby quantitatively revealing integrated performance differences and weaknesses among models.

%     \item \textbf{Quantitative analysis of abstention tendencies:}
%     The experiments demonstrated that LIT-RAGBench can quantitatively capture variations in answer-refusal tendencies. Improvements in abstention behavior can enhance response reliability but may also reduce accuracy on answerable questions.

%     \item \textbf{Implications for model improvement:}
%     These findings suggest that LLMs can be further improved through additional training and optimization of prompting strategies to better regulate abstention tendencies.
% \end{itemize}

%%%%%%%%%%%%%%%%%%%%%%%%%%%%%%%%%%%%%%%%%%%%%%%%%%%%%%%%%
%%%%%%%%%%%%%%%%%%%%%  Limitations  %%%%%%%%%%%%%%%%%%%%%
%%%%%%%%%%%%%%%%%%%%%%%%%%%%%%%%%%%%%%%%%%%%%%%%%%%%%%%%%
\section{Limitations}
Our main limitations are the small sample size and the imbalance across aspects. 
We designed an evaluation framework targeting realistic failure cases of $\mathcal{G}$, which have often been overlooked in previous studies, and built the accompanying dataset through careful human curation.
This process yielded a compact, high-quality dataset that covers the minimum occurrence patterns for each evaluation aspect, but it remains smaller than many existing benchmarks. 
While maintaining human-verified evaluation quality is important~\citep{maheshwari2024efficacysyntheticdatabenchmark,gill2025lostsyntheticevaluation}, expanding the benchmark to be more diverse and larger in scale is an essential direction for future work.

\section{Conclusion}
We constructed LIT-RAGBench, a benchmark comprising five evaluation categories designed from practical failure cases in real-world RAG systems.
Experiments on major LLMs showed that no model exceeded 90\% overall accuracy, and performance gaps were observed across categories.
These findings demonstrate that LIT-RAGBench provides a systematic and interpretable framework for assessing generator strengths and weaknesses in RAG.
For future work, we plan to extend the dataset and advance towards Agentic RAG~\citep{agentic-rag-survey}, where LLMs autonomously plan retrieval and reasoning steps.
We will release LIT-RAGBench as open source to support the advancement of RAG research and foster more reliable evaluation of large language models across diverse application domains.

%%%%%%%%%%%%%%%%%%%%%  Acknowledgement  %%%%%%%%%%%%%%%%%%%%%%
\section*{Acknowledgments}
We thank the anonymous reviewers for their constructive feedback.
We also thank neoAI Inc. for supporting this work by providing computational resources and covering experimental costs.
We are grateful to Koshiro Terasawa, Yuji Mochizuki, and Ryo Yagi, for their valuable assistance and insightful advice in constructing this benchmark.

%%%%%%%%%%%%%%%%%%%%%%%%%%%%%%%%%%%%%%%%%%%%%%%%%%%%%%%%%%%%%%%%%%%%%%%%%
%%%%%%%%%%%%%%%%%%%%%  Bibliographical References  %%%%%%%%%%%%%%%%%%%%%%
%%%%%%%%%%%%%%%%%%%%%%%%%%%%%%%%%%%%%%%%%%%%%%%%%%%%%%%%%%%%%%%%%%%%%%%%%
\clearpage
\section*{Bibliographical References}
\bibliographystyle{lrec2026-natbib}
\bibliography{references}

\end{document}